\documentclass[%
 aip,
 amsmath,amssymb,
 reprint,%
]{revtex4-1}

\usepackage{graphicx}
\usepackage{dcolumn}
\usepackage{bm}
\usepackage{siunitx}
\usepackage{amsmath}
\usepackage[utf8]{inputenc}
\usepackage[T1]{fontenc}
\usepackage{mathptmx}
\usepackage{etoolbox}
\usepackage{xcolor}
\usepackage{booktabs}

\makeatletter
\def\@email#1#2{%
 \endgroup
 \patchcmd{\titleblock@produce}
  {\frontmatter@RRAPformat}
  {\frontmatter@RRAPformat{\produce@RRAP{*#1\href{mailto:#2}{#2}}}\frontmatter@RRAPformat}
  {}{}
}%
\makeatother
\begin{document}

\preprint{AIP/123-QED}

\title[DiffCoder]{Coupled Diffusion–Encoder Models for Reconstruction of Flow Fields}
\author{AmirPouya Hemmasian}
 \email{Hemmas@cmu.edu}
 \affiliation{Mechanical Engineering Department, Carnegie Mellon University, Pittsburgh, PA, USA}
 
\author{Amir Barati Farimani}
 \email{Barati@cmu.edu}
 \affiliation{Mechanical Engineering Department, Carnegie Mellon University, Pittsburgh, PA, USA}
 \affiliation{Machine Learning Department, Carnegie Mellon University, Pittsburgh, PA, USA}

\date{\today}

\begin{abstract}
Data-driven flow-field reconstruction typically relies on autoencoder architectures that compress high-dimensional states into low-dimensional latent representations. However, classical approaches such as variational autoencoders (VAEs) often struggle to preserve the higher-order statistical structure of fluid flows when subjected to strong compression. We propose DiffCoder, a coupled framework that integrates a probabilistic diffusion model with a conventional convolutional ResNet encoder and trains both components end-to-end. The encoder compresses the flow field into a latent representation, while the diffusion model learns a generative prior over reconstructions conditioned on the compressed state. This design allows DiffCoder to recover distributional and spectral properties that are not strictly required for minimizing pointwise reconstruction loss but are critical for faithfully representing statistical properties of the flow field. We evaluate DiffCoder and VAE baselines across multiple model sizes and compression ratios on a challenging dataset of Kolmogorov flow fields. Under aggressive compression, DiffCoder significantly improves the spectral accuracy while VAEs exhibit substantial degradation. Although both methods show comparable relative L2 reconstruction error, DiffCoder better preserves the underlying distributional structure of the flow. At moderate compression levels, sufficiently large VAEs remain competitive, suggesting that diffusion-based priors provide the greatest benefit when information bottlenecks are severe. These results demonstrate that the generative decoding by diffusion offers a promising path toward compact, statistically consistent representations of complex flow fields.
\end{abstract}

\maketitle

\section{Introduction} 
\label{sec:intro}

Complex fluid flows are governed by high-dimensional, nonlinear dynamics that are expensive to simulate, store, and analyze. 
In many scientific and engineering settings, especially where real-time control or embedded deployment is required, there is a growing need for compressing these flows into compact representations that still preserve their spatial and statistical structure. 
Traditional approaches such as Proper Orthogonal Decomposition (POD) and dynamic mode decomposition (DMD) often fail to capture the nonlinear multiscale dynamics present in chaotic regimes \cite{Fukami2021ModelOrderReductionNN}.

Deep learning-based autoencoders have shown significant promise for data-driven flow compression and reconstruction, offering the ability to learn nonlinear mappings from high-dimensional states into latent representations. 
Variational Autoencoders (VAEs), in particular, enable sampling and probabilistic modeling, but are known to produce blurry or smoothed reconstructions under aggressive compression \cite{SoleraRico2024BetaVAETransformers, Glaws2020DeepLearningInSitu, Racca2023PredictingTurbulentDynamics}. 
This blurriness is largely due to the limitations of the Gaussian prior and the variational bound used in training, which may fail to capture the full complexity of the flow field's data distribution.

Recent developments in generative modeling, particularly score-based diffusion models, offer a powerful alternative for learning complex distributions through a stochastic denoising process \cite{BayesianCondDiffusionTurbulence2023, UnfoldingTimeGenerativeTurbulentFlows2024}. 
These models have demonstrated strong performance across applications in computer vision, physics simulation, and design. Notably, recent advances have applied diffusion models to flow field super-resolution \cite{shu2023physics}, stress estimation in solids \cite{jadhav2023stressd}, and melt pool modeling in additive manufacturing \cite{ogoke2024inexpensive}, highlighting their ability to model complex physical phenomena with high fidelity. 
Extensions of diffusion models for inverse design \cite{jadhav2024generative}, uncertainty quantification \cite{shu2024zero}, and text-based simulations \cite{zhou2410text2pde} further demonstrate their versatility in modeling structured physical data. 
These developments support the use of diffusion models as expressive generative priors capable of capturing the underlying distribution of quantities in physical systems.

In this work, we propose \textbf{DiffCoder}, a hybrid framework that couples a convolutional ResNet encoder with a conditional diffusion model. 
The encoder compresses the flow field into a latent code (compressed field) and the diffusion model learns a generative prior over the space of possible reconstructions, conditioned on the compressed field. 
This setup enables a recovery mechanism that preserves high-order statistics and spatial detail, even at significant compression rates.

Several recent studies have explored alternatives to conventional CAEs for flow modeling, including vector-quantized autoencoders, transformer-based reduced-order models, and physics-informed loss design \cite{Momenifar2022AutoencoderTurbulence, SoleraRico2024BetaVAETransformers, Glaws2020DeepLearningInSitu}. 
DiffCoder is most closely related to conditional diffusion applied to physical systems \cite{BayesianCondDiffusionTurbulence2023, shu2023physics}, but differs by embedding a learned representation from an encoder network that is trained together as part of the diffusion training procedure. Additional work has demonstrated the effectiveness of convolutional architectures for learning spatiotemporal generative priors \cite{Racca2023PredictingTurbulentDynamics, UnfoldingTimeGenerativeTurbulentFlows2024} and integrating diffusion-based generative models into physical modeling pipelines \cite{ogoke2025deep, shu2024zero}. 
Our approach builds on these directions and contributes a unified system trained end-to-end.

The remainder of the paper is organized as follows: Section~\ref{sec:related_work} reviews previous work in flow compression and generative modeling. Section~\ref{sec:diffusion_models} provides a brief background on diffusion models. Section~\ref{sec:method} details the architecture of DiffCoder and its training and inference procedures. In Section~\ref{sec:experiments}, we present results comparing DiffCoder to a Variational Autoencoder baseline built with the same building blocks and parameter count, followed by a discussion and conclusion.

\section{Related Works}
\label{sec:related_work}

\subsection{Reduced-Order Modeling and Flow Compression}

Classical reduced-order modeling techniques project high-dimensional flow states onto low-dimensional subspaces. Proper Orthogonal Decomposition (POD) and its variants have been foundational for flow analysis and reconstruction \cite{sirovich1987turbulence, bui2004aerodynamic}, with extensions like Gappy-POD enabling reconstruction from incomplete data \cite{venturi2004gappy}. Sparse reduced-order modeling has connected sensor-based dynamics to full-state estimation \cite{loiseau2018sparse}. However, linear methods struggle to capture the nonlinear, multiscale dynamics of complex and chaotic fluid flows \cite{Fukami2021ModelOrderReductionNN}. 

Deep learning-based autoencoders address this limitation by learning nonlinear mappings between high-dimensional states and compact latent representations. Erichson et al.\ \cite{erichson2020shallow} demonstrated that shallow neural networks can reconstruct flow fields from limited sensor measurements, while Fukami et al.\ \cite{fukami2021global} combined Voronoi tessellation with deep learning for global field reconstruction from sparse sensors. Extensions using $\beta$-VAEs \cite{SoleraRico2024BetaVAETransformers} and physics-constrained architectures \cite{Glaws2020DeepLearningInSitu} have improved generalization, though VAE-based decoders often produce overly smooth reconstructions under aggressive compression due to limitations of the Gaussian likelihood assumption.

\subsection{Generative Models for Flow Super-Resolution}

Generative adversarial networks (GANs) have been applied to fluid flow super-resolution with notable success. G\"uemes et al.\ \cite{guemes2022super} developed super-resolution GANs for randomly-seeded fields, achieving high-quality reconstructions in experimental settings. Physics-informed GANs have been used for turbulence enrichment \cite{super_3d_gan_1}, and unsupervised deep learning approaches have shown promise for super-resolution reconstruction of turbulence \cite{super_3d_gan_2}. However, GAN training instabilities and mode collapse remain persistent challenges. For image super-resolution, deep convolutional networks \cite{dong2015image} and iterative refinement approaches \cite{saharia2022image} have established strong baselines that have been adapted for scientific applications.

\subsection{Diffusion Models for Physical Systems}

Diffusion models have emerged as powerful generative priors, surpassing GANs in image synthesis quality \cite{dhariwal2021diffusion}. Ho et al.\ \cite{ho2020denoising} established the foundational DDPM framework, while Song et al.\ \cite{song2020denoising} introduced DDIM for accelerated deterministic sampling. Improved training strategies \cite{nichol2021improved} and efficient loss weighting schemes based on signal-to-noise ratio \cite{hang2023efficient} have enhanced sample quality and training stability.

In the context of fluid mechanics, Shu et al.\ \cite{shu2023physics} developed physics-informed diffusion models for high-fidelity flow field reconstruction, demonstrating superior spectral accuracy compared to deterministic baselines. Liu and Thuerey \cite{liu2024uncertainty} applied diffusion models as uncertainty-aware surrogates for airfoil flow simulations. Yang and Sommer \cite{yang2023denoising} explored diffusion for fluid field prediction, Li et al.\ \cite{li2023multi} reconstructed turbulent rotating flows using generative diffusion, and Abaidi and Adams \cite{abaidi2025exploring} investigated diffusion models for compressible fluid field prediction.

Physics-informed approaches have been integrated into diffusion training: Bastek et al.\ \cite{bastek2024physics} developed physics-informed diffusion models, while Shan et al.\ \cite{shan2024pird} proposed physics-informed residual diffusion for flow field reconstruction. Lu and Xu \cite{lu2024generative} introduced generative downscaling of PDE solvers with physics-guided diffusion.

\section{Background: Diffusion Models}
\label{sec:diffusion_models}
Diffusion models are a class of generative models that learn to reverse a gradual noising process. Given data $\mathbf{x}_0$ drawn from a target distribution, the forward process progressively corrupts the data over $T$ timesteps by adding Gaussian noise:
\begin{equation}
q(\mathbf{x}_t | \mathbf{x}_{t-1}) = \mathcal{N}(\mathbf{x}_t; \sqrt{1-\beta_t}\, \mathbf{x}_{t-1}, \beta_t \mathbf{I}),
\end{equation}
where $\{\beta_t\}_{t=1}^T$ is a variance schedule controlling the noise magnitude at each step. Common choices include linear schedules \cite{ho2020denoising} and cosine schedules \cite{nichol2021improved}. An alternative parameterization defines the schedule through the log signal-to-noise ratio $\lambda_t = \log(\bar{\alpha}_t / (1 - \bar{\alpha}_t))$, from which $\bar{\alpha}_t = \sigma(\lambda_t)$ where $\sigma$ denotes the sigmoid function.

Defining $\alpha_t = 1 - \beta_t$ and $\bar{\alpha}_t = \prod_{s=1}^t \alpha_s$, the marginal distribution at any timestep $t$ can be expressed in closed form:
\begin{equation}
q(\mathbf{x}_t | \mathbf{x}_0) = \mathcal{N}(\mathbf{x}_t; \sqrt{\bar{\alpha}_t}\, \mathbf{x}_0, (1-\bar{\alpha}_t) \mathbf{I}).
\end{equation}
This allows direct sampling of $\mathbf{x}_t$ from $\mathbf{x}_0$ via the reparameterization $\mathbf{x}_t = \sqrt{\bar{\alpha}_t}\, \mathbf{x}_0 + \sqrt{1-\bar{\alpha}_t}\, \boldsymbol{\epsilon}$, where $\boldsymbol{\epsilon} \sim \mathcal{N}(\mathbf{0}, \mathbf{I})$.

Generation proceeds by learning the reverse process $p_\theta(\mathbf{x}_{t-1} | \mathbf{x}_t)$, which iteratively denoises samples starting from Gaussian noise $\mathbf{x}_T \sim \mathcal{N}(\mathbf{0}, \mathbf{I})$. The reverse process is parameterized as a Gaussian with learned mean:
\begin{equation}
p_\theta(\mathbf{x}_{t-1} | \mathbf{x}_t) = \mathcal{N}(\mathbf{x}_{t-1}; \boldsymbol{\mu}_\theta(\mathbf{x}_t, t), \sigma_t^2 \mathbf{I}).
\end{equation}
Training minimizes a variational bound that reduces to a denoising objective. The network can be trained to predict the noise $\boldsymbol{\epsilon}$, the clean data $\mathbf{x}_0$, or the velocity $\mathbf{v}_t = \sqrt{\bar{\alpha}_t}\, \boldsymbol{\epsilon} - \sqrt{1-\bar{\alpha}_t}\, \mathbf{x}_0$. The velocity parameterization, introduced by Salimans and Ho \cite{salimans2022progressive}, provides numerical stability across all noise levels and is particularly well-suited for few-step sampling.

Denoising Diffusion Implicit Models (DDIM) \cite{song2020denoising} reformulate the reverse process as a deterministic mapping, enabling sampling with far fewer steps than the original DDPM formulation \cite{ho2020denoising}. Given a predicted estimate $\hat{\mathbf{x}}_0$ at timestep $t$, DDIM computes:
\begin{equation}
\mathbf{x}_{t-1} = \sqrt{\bar{\alpha}_{t-1}}\, \hat{\mathbf{x}}_0 + \sqrt{1-\bar{\alpha}_{t-1}}\, \frac{\mathbf{x}_t - \sqrt{\bar{\alpha}_t}\, \hat{\mathbf{x}}_0}{\sqrt{1-\bar{\alpha}_t}}.
\label{eq:ddim}
\end{equation}
This deterministic formulation permits skipping timesteps, reducing sampling from hundreds of steps to tens while maintaining generation quality.

Conditional diffusion models extend this framework by incorporating auxiliary information $\mathbf{c}$ into the denoising network, yielding $\boldsymbol{\epsilon}_\theta(\mathbf{x}_t, t, \mathbf{c})$ or equivalently $\mathbf{v}_\theta(\mathbf{x}_t, t, \mathbf{c})$. Common conditioning mechanisms include concatenation along the channel dimension, cross-attention, or adaptive normalization layers. In our setting, the conditioning signal $\mathbf{c} = \mathbf{z}$ is the compressed representation produced by a deterministic encoder, enabling the diffusion model to reconstruct high-fidelity flow fields accordingly.

\section{Methodology}
\label{sec:method}

Building on the premise that diffusion models provide richer generative priors than Gaussian-based decoders, \textbf{DiffCoder} operationalizes this concept through a two-stage design: a deterministic encoder that compresses the flow field and a conditional diffusion decoder that reconstructs the field via iterative denoising. The following subsections describe the architecture, training and inference, and baseline configuration.

\subsection{Architecture Overview}

\begin{figure*}[htbp]
\centering
\includegraphics[width=\textwidth]{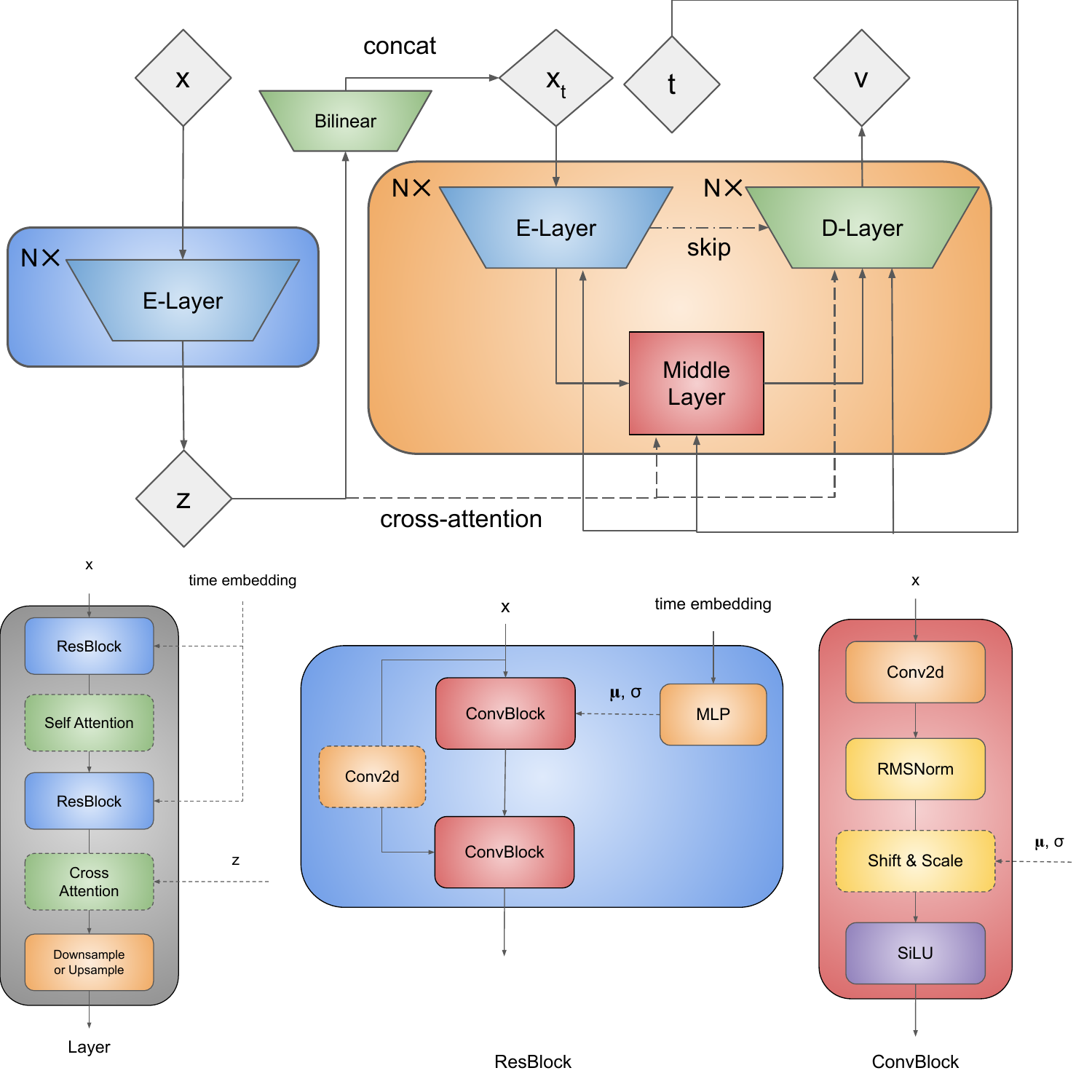}
\caption{Top: DiffCoder's equivalent of a decoder is a conditional U-Net that performs iterative denoising in the reverse diffusion process. The encoder produces $\mathbf{z}$, which conditions the U-Net via concatenation to the input and optionally cross-attention. The U-Net receives timestep $t$ through shift-and-scale modulation and predicts the target (in our case velocity $\mathbf{v}$) for the reverse diffusion process. Bottom: The building blocks of DiffCoder and the VAE baseline.}
\label{fig:architecture}
\end{figure*}

Comprising convolutional and residual building blocks, the encoder and conditional denoiser of DiffCoder allow it to learn the best encoding for the generative reconstruction process (i.e. reverse diffusion). Figure~\ref{fig:architecture} illustrates the overall architecture (top) and the building blocks (bottom).

In the VAE baseline, the encoder compresses the input field $\mathbf{x}$ into a latent representation $\mathbf{z}$, and a symmetric decoder reconstructs $\hat{\mathbf{x}}$ directly. In DiffCoder, the same encoder produces $\mathbf{z}$, but the reconstruction is performed by a U-Net operating within a reverse diffusion process. The U-Net takes as input the noisy field $\mathbf{x}_t$, the timestep $t$ (injected via shift-and-scale modulation), and the conditioning signal $\mathbf{z}$ (provided through concatenation and optionally cross-attention). Rather than directly outputting the reconstruction, the U-Net predicts the velocity $\mathbf{v}$ used in the reverse diffusion process, iteratively refining the output over multiple denoising steps.

The fundamental building block is a Convolution Block consisting of a Conv2d layer, RMSNorm, and SiLU activation. When time conditioning is required (in the U-Net only), the block incorporates adaptive shift-and-scale modulation derived from the timestep embedding via a linear projection. ResNet Blocks combine two Convolution Blocks with a residual connection. Each Layer contains ResNet Blocks followed by optional self-attention and cross-attention modules, then a downsample or upsample operation; cross-attention is used exclusively in the U-Net for conditioning on the compressed representation.

The \textbf{encoder} consists of a sequence of encoder layers (E-Layers), each comprising two ResNet Blocks followed by a downsampling operation (except the final layer). This progressively compresses the input field $\mathbf{x} \in \mathbb{R}^{H \times W}$ to a compressed representation $\mathbf{z} \in \mathbb{R}^{h \times w}$ where $h < H$ and $w < W$. Adjustments can be made according to the properties of the problem and the dataset at hand, such as circular padding for periodic boundary conditions. The VAE \textbf{decoder} mirrors the encoder in reverse: a sequence of decoder layers (D-Layers) with two ResNet Blocks and an upsampling operation.

The \textbf{U-Net} follows the standard encoder-decoder structure with skip connections \cite{ronneberger2015u}. The contracting path (E-Layers) downsamples through successive layers, a middle layer processes the bottleneck features, and the expanding path (D-Layers) upsamples with skip connections from corresponding encoder layers. The U-Net incorporates timestep embeddings that modulate the Convolution Blocks via shift-and-scale operations, and receives the compressed field $\mathbf{z}$ through concatenation with the input and optionally cross attention. Empirically, we observe that models without attention modules outperform attention-based variants in reconstruction quality for this application. This finding is consistent with the predominantly local spatial correlations in flow fields, which favor convolutional operations over global attention mechanisms.

\subsection{Training and Inference}

Training follows the discrete-time diffusion framework described in Section~\ref{sec:diffusion_models} with $T = 1000$ timesteps. We adopt a sigmoid noise schedule \cite{chen2023importance, song2020score}, which defines the log signal-to-noise ratio as a linear interpolation:
\begin{equation}
\lambda_t = (1 - t/T) \lambda_{\min} + (t/T) \lambda_{\max},
\end{equation}
where $\lambda_{\min}$ and $\lambda_{\max}$ are hyperparameters controlling the noise range, and $\bar{\alpha}_t = \sigma(\lambda_t)$. This schedule allocates more capacity to intermediate noise levels compared to linear alternatives.

For each training sample, a timestep $t \in [1, T]$ is drawn uniformly, and the network predicts the velocity target $\mathbf{v}_t$ conditioned on the latent representation $\mathbf{z}$ from the encoder:
\begin{equation}
\hat{\mathbf{v}}_t = \mathbf{v}_\theta(\mathbf{x}_t, t, \mathbf{z}).
\end{equation}
Among the three common prediction targets ($\boldsymbol{\epsilon}$, $\mathbf{x}_0$, and $\mathbf{v}$), the velocity objective produces the most stable optimization and accurate reconstructions in our experiments \cite{salimans2022progressive}. The training loss is:
\begin{equation}
\mathcal{L} = \mathbb{E}_{\mathbf{x}_0, t, \boldsymbol{\epsilon}}\left[ w_t \left\| \mathbf{v}_\theta(\mathbf{x}_t, t, \mathbf{z}) - \mathbf{v}_t \right\|_2^2 \right],
\end{equation}
where $w_t = \max(\text{SNR}(t), 1)$ with $\text{SNR}(t) = \bar{\alpha}_t / (1 - \bar{\alpha}_t)$ is a truncated signal-to-noise ratio weight \cite{salimans2022progressive} that emphasizes high-noise timesteps. The encoder and U-Net are optimized jointly using AdamW with a OneCycle learning rate schedule.

At inference time, reconstruction proceeds via DDIM sampling (Eq.~\ref{eq:ddim}). Starting from Gaussian noise $\mathbf{x}_T \sim \mathcal{N}(\mathbf{0}, \mathbf{I})$, the model iteratively denoises while conditioning on the compressed field $\mathbf{z}$ at each step, transforming random noise into a physically plausible flow field. We use 20 DDIM steps for all experiments.

To isolate the contribution of the diffusion-based decoder, we compare against a VAE baseline with matched encoder architecture. The VAE uses the same encoder and a symmetric decoder (as described above), trained with the standard evidence lower bound (ELBO) objective combining reconstruction loss and KL divergence regularization. This comparison ensures that performance differences arise from the generative modeling approach rather than architectural capacity.

\section{Experiments}
\label{sec:experiments}

We evaluate DiffCoder on two-dimensional Kolmogorov flow at $\text{Re} = 1000$, comparing against a VAE baseline across varying compression ratios and model capacities.

\subsection{Dataset and Setup}

We use the Kolmogorov flow dataset from the work of Li et al.\cite{li2023scalable} which consists of vorticity field snapshots from 120 Kolmogorov flow trajectories simulated at Reynolds number $\text{Re} = 1000$ on a $256 \times 256$ grid with periodic boundary conditions. We use 100 trajectories for training and 20 for testing.

The VAE baseline uses an identical encoder architecture to DiffCoder, with a symmetric convolutional decoder. Both models are trained with the AdamW optimizer using a OneCycle learning rate schedule with maximum learning rate $5 \times 10^{-4}$, batch size 16, for 10 epochs. At inference, DiffCoder uses 20 DDIM steps.

\subsection{Evaluation Metrics}

Since we want to evaluate the spectral accuracy of the model rather than only the point-wise accuracy, we define two additional metrics to assess the results, leading to three total metrics as follows:

\paragraph{Relative $L_2$ error (vorticity).} We compute the relative $L_2$ error between reconstructed and ground-truth vorticity fields:
\begin{equation}
\epsilon_\omega = \frac{\| \omega_{\text{rec}} - \omega_{\text{gt}} \|_2}{\| \omega_{\text{gt}} \|_2},
\end{equation}
where fields are flattened to vectors for the norm computation.

\paragraph{Spectral error (full).} To evaluate preservation of the energy spectrum, we compute the relative $L_2$ error of the log kinetic energy spectrum:
\begin{equation}
\epsilon_E = \frac{\| \log E_{\text{rec}}(k) - \log E_{\text{gt}}(k) \|_2}{\| \log E_{\text{gt}}(k) \|_2},
\end{equation}
where $E(k)$ is the kinetic energy spectrum as a function of wavenumber $k$.

\paragraph{High-wavenumber spectral error.} Since small-scale structures are most challenging to reconstruct, we also report the spectral error restricted to the highest 25\% of wavenumbers, denoted $\epsilon_E^{\text{high}}$.

\subsection{Results}

\begin{figure}[htbp]
\centering
\includegraphics[width=\columnwidth]{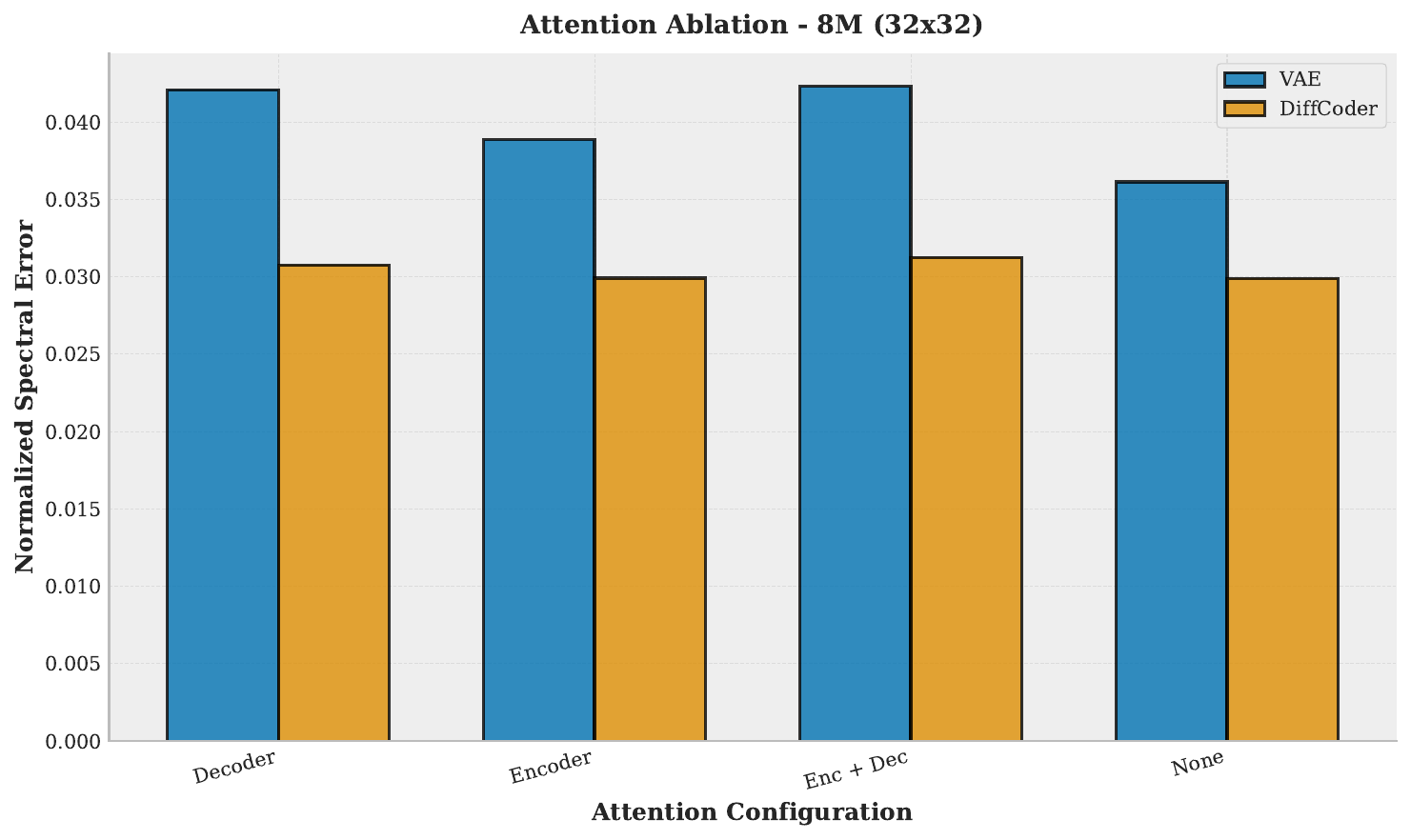}
\caption{Attention mechanism ablation study for 8M parameter scale and encoder depth 3 (64$\times$ compression). Comparison of VAE and DiffCoder across four attention configurations: None, Encoder only, Decoder/U-Net only, and Encoder+Decoder/U-Net. Results show the impact of different attention mechanisms on normalized spectral error (lower is better).}
\label{fig:attention_ablation}
\end{figure}

First, we conducted a series of experiments with the 8M parameter scale with encoder depth of 3 (compressing to $32 \times 32$ fields, which means $64\times$ compression) to determine whether the attention mechanism benefits DiffCoder or the baseline VAE. 
 We evaluated four configurations: no attention in either encoder or decoder (or U-Net), attention in encoder only,  attention in decoder (or U-Net) only, and attention in both components. Removing attention entirely yields the best performance for DiffCoder, likely because the relatively small spatial dimensions and local nature of flow structures do not benefit from global attention mechanisms. We also compared linear, cosine, and sigmoid noise schedules. The sigmoid schedule provides the best results, offering smoother transitions in signal-to-noise ratio during the diffusion process. Based on these ablations, all subsequent experiments use DiffCoder without attention and with a sigmoid noise schedule.improves model performance. Figure~\ref{fig:attention_ablation} summarizes the results of the four attention configurations for both models. 

\begin{figure}[htbp]
\centering
\includegraphics[width=\columnwidth]{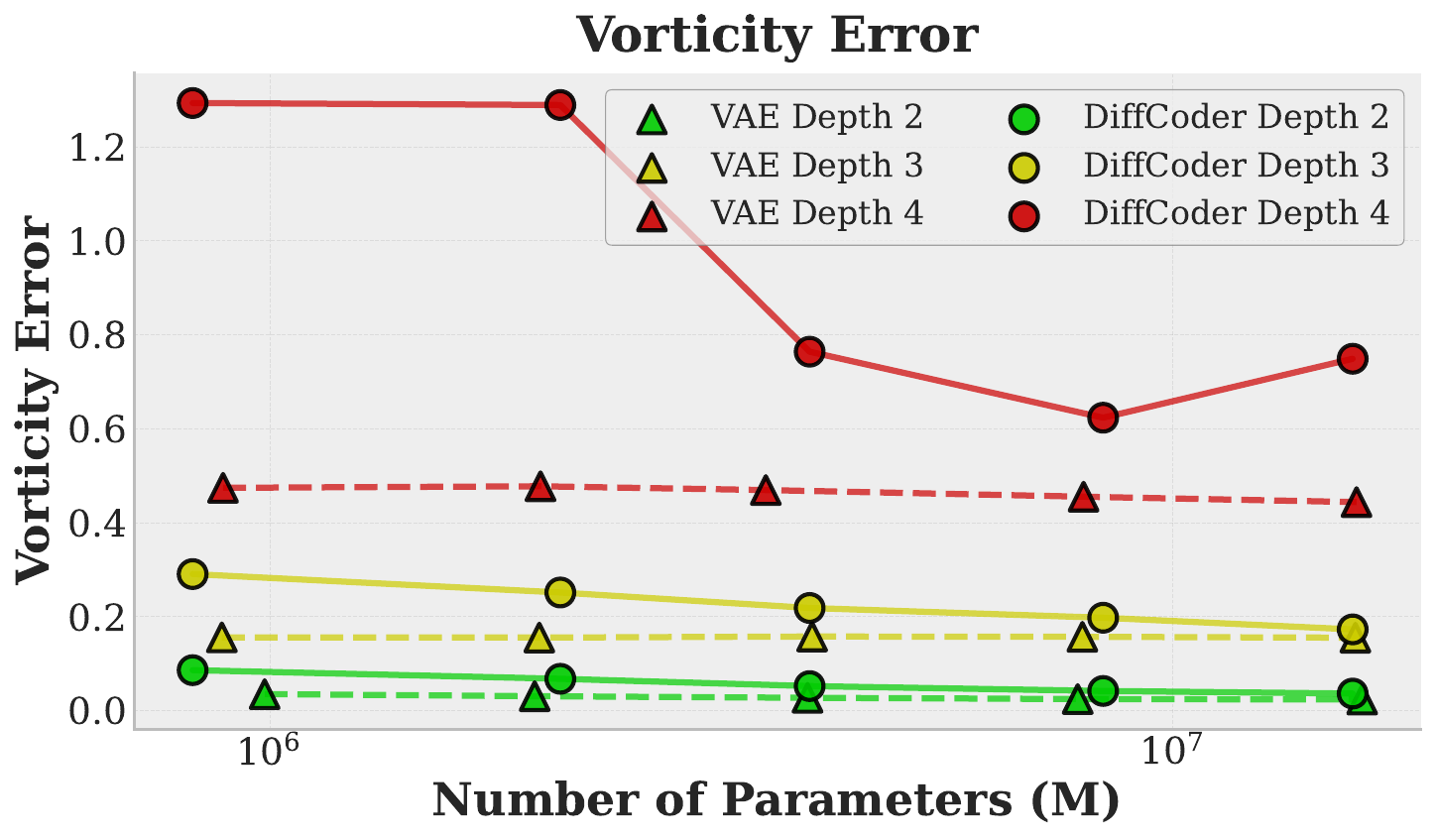}
\caption[DiffCoder: vorticity errors]{Vorticity error across model sizes and encoder depths. Both models slightly improve with increased capacity, with the VAE being the consistent winner. In deeper compressions, both models fail to achieve an acceptable performance with VAE not even reaching 40\% and DiffCoder showing even worse performance.}
\label{fig:vorticity}
\end{figure}

\begin{figure}[htbp]
\centering
\includegraphics[width=\columnwidth]{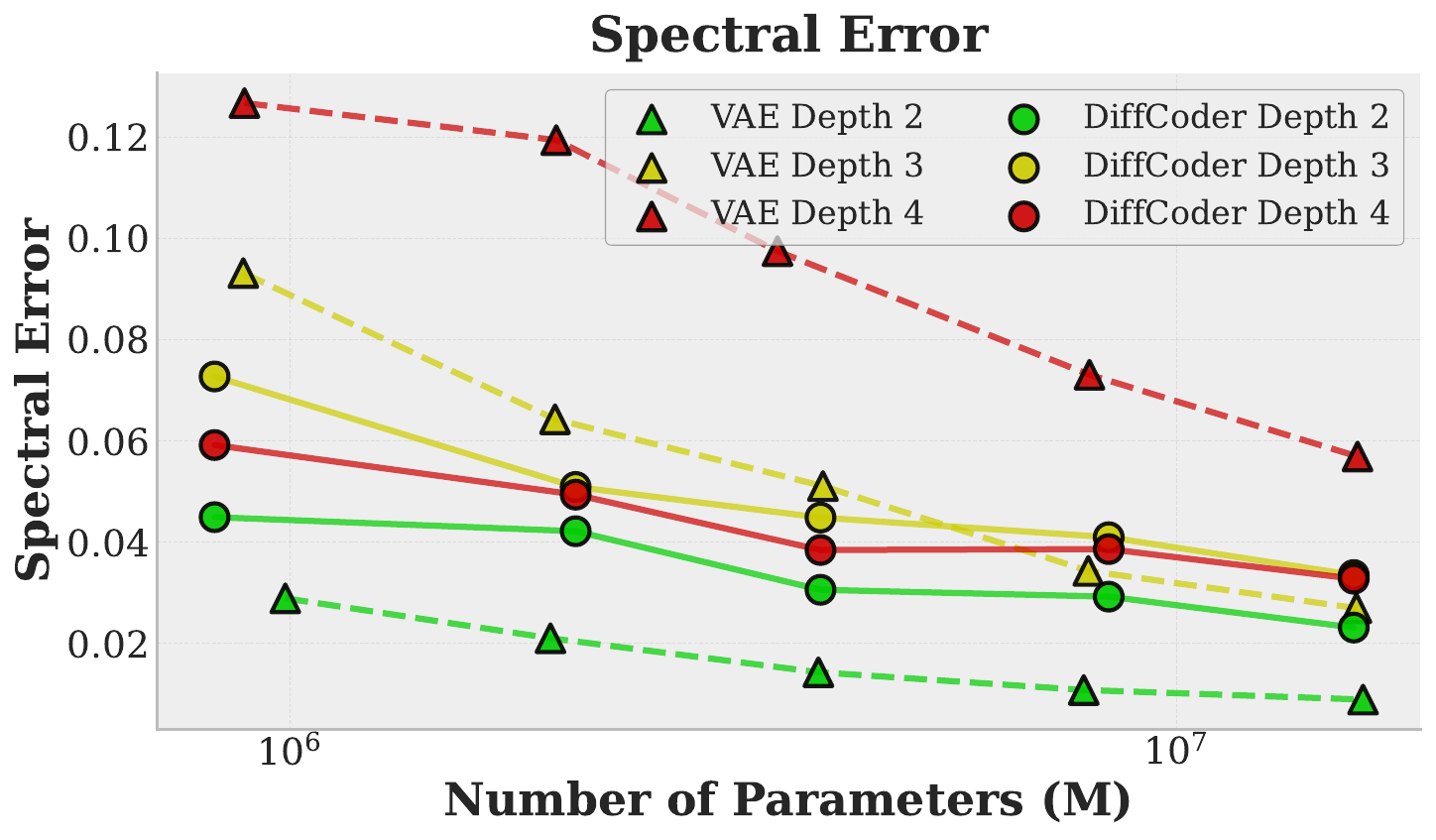}
\caption[DiffCoder: spectral errors]{Spectral error across model sizes and encoder depths. DiffCoder demonstrates substantial advantages at encoder depth 4, where spectral content is lost the most and are the most difficult to preserve. At shallow depths, both approaches perform comparably, with DiffCoder outperforming VAE only at depth 3 for smaller model sizes, while underperforming VAE in the remainder of the experiments.}
\label{fig:spectral}
\end{figure}

\begin{figure}[htbp]
\centering
\includegraphics[width=\columnwidth]{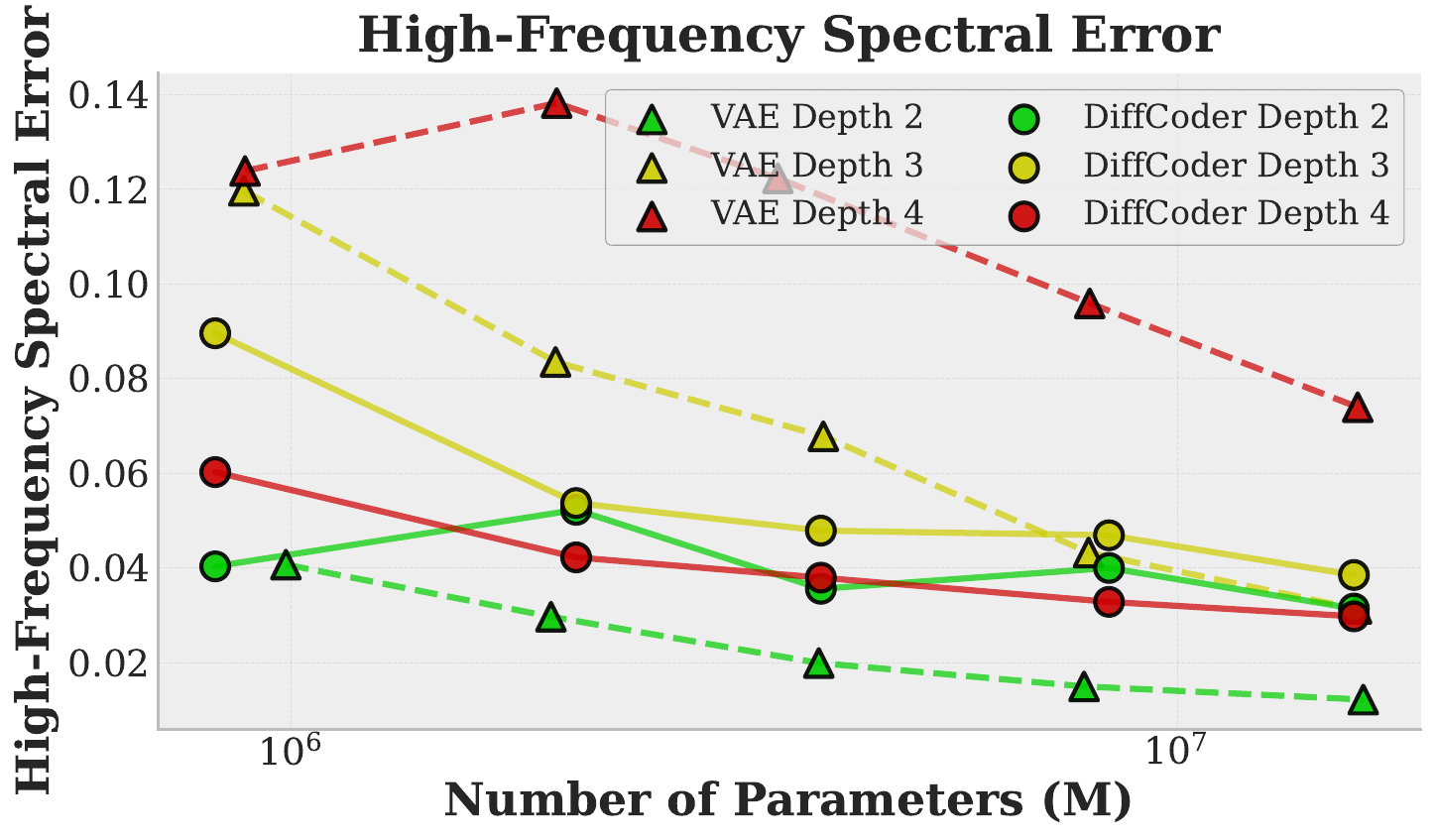}
\caption[DiffCoder: high-frequency spectral errors]{High-frequency spectral error across model sizes and encoder depths. The advantage of DiffCoder is most pronounced at depth 4 where high-wavenumber content are almost if not totally lost. This metric isolates the ability to recover small-scale structures under deep compression.}
\label{fig:highfreq_spectral}
\end{figure}

Moving on to the main experiments, 
We train both DiffCoder and the VAE baseline at five model scales: 1M, 2M, 4M, 8M, and 16M parameters, with each component (encoder, decoder/U-Net) scaled proportionally. We evaluate three encoder depth settings: depth 2, depth 3, and depth 4, corresponding to progressively deeper compressions. Each additional depth level increases the spatial downsampling factor by a power of 2 across each axis. Figures~\ref{fig:vorticity}--\ref{fig:highfreq_spectral} present the three evaluation metrics across encoder depths and model sizes, and tables~\ref{tab:gain_vorticity}--\ref{tab:gain_highfreq_spectral} summarize the relative performance of DiffCoder compared to the VAE baseline across all configurations. We will take a closer look at each compression regime next.

\paragraph{Mild compression (encoder depth 2).} When compression is minimal, the VAE outperforms DiffCoder on both vorticity and spectral metrics across all model sizes. The simpler decoder of the VAE suffices when the latent representation retains most of the original information. The high-frequency spectral error (Figure~\ref{fig:highfreq_spectral}) shows similar trends as both models preserve fine-scale content reasonably well, with VAE holding a visible advantage.

\paragraph{Moderate compression (encoder depth 3).} At this intermediate compression level, the two approaches achieve comparable performance, with DiffCoder showing advantages in spectral fidelity at smaller model scales. The crossover becomes visible in Figures~\ref{fig:spectral} and~\ref{fig:highfreq_spectral}, where VAE curves approach or intersect DiffCoder in spectral error as model size increases, which means that VAE can also excel in matching the spectrum given enough parameters.

\paragraph{Deep compression (encoder depth 4).} Under aggressive compression, DiffCoder demonstrates clear superiority in spectral metrics. While both models exhibit elevated vorticity reconstruction error (Figure~\ref{fig:vorticity}), reflecting the fundamental information loss, DiffCoder substantially outperforms the VAE in spectral error (Figure~\ref{fig:spectral}). This advantage is most pronounced for the high-wavenumber spectral error $\epsilon_E^{\text{high}}$ (Figure~\ref{fig:highfreq_spectral}), indicating that DiffCoder better recovers small-scale turbulent structures even when pointwise reconstruction is challenging.

This trend confirms our central hypothesis: diffusion-based decoding provides the greatest benefit when the compression bottleneck forces the decoder to generate plausible high-frequency content rather than simply reconstruct it from latent information.


\begin{table*}[!ht]
\centering
\caption[DiffCoder: gain in vorticity error]{Relative vorticity error: DiffCoder vs VAE. Values show $\varepsilon^\text{VAE}_\omega \rightarrow \varepsilon^\text{Diff}_\omega$ with percentage change in parentheses. Negative values (green) indicate DiffCoder outperforms VAE.}
\label{tab:gain_vorticity}

\begin{tabular}{lccc}
\toprule
Model Size & Depth 2 & Depth 3 & Depth 4 \\
\midrule
1M & 0.0353 $\rightarrow$ 0.0863 (\textcolor{red!70!black}{+144.8\%}) & 0.1561 $\rightarrow$ 0.2908 (\textcolor{red!70!black}{+86.3\%}) & 0.4744 $\rightarrow$ 1.2933 (\textcolor{red!70!black}{+172.6\%}) \\
2M & 0.0309 $\rightarrow$ 0.0679 (\textcolor{red!70!black}{+120.0\%}) & 0.1557 $\rightarrow$ 0.2516 (\textcolor{red!70!black}{+61.6\%}) & 0.4777 $\rightarrow$ 1.2891 (\textcolor{red!70!black}{+169.9\%}) \\
4M & 0.0275 $\rightarrow$ 0.0524 (\textcolor{red!70!black}{+90.8\%}) & 0.1579 $\rightarrow$ 0.2187 (\textcolor{red!70!black}{+38.4\%}) & 0.4696 $\rightarrow$ 0.7641 (\textcolor{red!70!black}{+62.7\%}) \\
8M & 0.0248 $\rightarrow$ 0.0421 (\textcolor{red!70!black}{+69.5\%}) & 0.1573 $\rightarrow$ 0.1979 (\textcolor{red!70!black}{+25.8\%}) & 0.4555 $\rightarrow$ 0.6236 (\textcolor{red!70!black}{+36.9\%}) \\
16M & 0.0238 $\rightarrow$ 0.0365 (\textcolor{red!70!black}{+53.8\%}) & 0.1555 $\rightarrow$ 0.1731 (\textcolor{red!70!black}{+11.3\%}) & 0.4440 $\rightarrow$ 0.7491 (\textcolor{red!70!black}{+68.7\%}) \\
\bottomrule
\end{tabular}

\end{table*}

\begin{table*}[!ht]
\centering
\caption[DiffCoder: gain in spectral error]{Relative spectral error: DiffCoder vs VAE. Values show $\varepsilon^\text{VAE}_\text{spec} \rightarrow \varepsilon^\text{Diff}_\text{spec}$ with percentage change in parentheses. Negative values (green) indicate DiffCoder outperforms VAE.}
\label{tab:gain_spectral}

\begin{tabular}{lccc}
\toprule
Model Size & Depth 2 & Depth 3 & Depth 4 \\
\midrule
1M & 0.0289 $\rightarrow$ 0.0450 (\textcolor{red!70!black}{+55.5\%}) & 0.0932 $\rightarrow$ 0.0728 (\textcolor{green!70!black}{-21.9\%}) & 0.1268 $\rightarrow$ 0.0592 (\textcolor{green!70!black}{-53.3\%}) \\
2M & 0.0210 $\rightarrow$ 0.0422 (\textcolor{red!70!black}{+100.5\%}) & 0.0643 $\rightarrow$ 0.0509 (\textcolor{green!70!black}{-20.8\%}) & 0.1195 $\rightarrow$ 0.0494 (\textcolor{green!70!black}{-58.6\%}) \\
4M & 0.0143 $\rightarrow$ 0.0306 (\textcolor{red!70!black}{+113.9\%}) & 0.0511 $\rightarrow$ 0.0449 (\textcolor{green!70!black}{-12.2\%}) & 0.0975 $\rightarrow$ 0.0385 (\textcolor{green!70!black}{-60.6\%}) \\
8M & 0.0108 $\rightarrow$ 0.0293 (\textcolor{red!70!black}{+170.6\%}) & 0.0343 $\rightarrow$ 0.0410 (\textcolor{red!70!black}{+19.4\%}) & 0.0731 $\rightarrow$ 0.0387 (\textcolor{green!70!black}{-47.1\%}) \\
16M & 0.0089 $\rightarrow$ 0.0231 (\textcolor{red!70!black}{+159.6\%}) & 0.0270 $\rightarrow$ 0.0335 (\textcolor{red!70!black}{+24.3\%}) & 0.0570 $\rightarrow$ 0.0328 (\textcolor{green!70!black}{-42.4\%}) \\
\bottomrule
\end{tabular}

\end{table*}

\begin{table*}[!ht]
\centering
\caption[DiffCoder: gain in high-frequency spectral error]{Relative high-frequency spectral error: DiffCoder vs VAE. Values show $\varepsilon^\text{VAE}_\text{HF} \rightarrow \varepsilon^\text{Diff}_\text{HF}$ with percentage change in parentheses. Negative values (green) indicate DiffCoder outperforms VAE.}
\label{tab:gain_highfreq_spectral}

\begin{tabular}{lccc}
\toprule
Model Size & Depth 2 & Depth 3 & Depth 4 \\
\midrule
1M & 0.0407 $\rightarrow$ 0.0403 (\textcolor{green!70!black}{-0.9\%}) & 0.1197 $\rightarrow$ 0.0896 (\textcolor{green!70!black}{-25.1\%}) & 0.1239 $\rightarrow$ 0.0602 (\textcolor{green!70!black}{-51.4\%}) \\
2M & 0.0297 $\rightarrow$ 0.0523 (\textcolor{red!70!black}{+76.0\%}) & 0.0835 $\rightarrow$ 0.0537 (\textcolor{green!70!black}{-35.7\%}) & 0.1382 $\rightarrow$ 0.0422 (\textcolor{green!70!black}{-69.5\%}) \\
4M & 0.0199 $\rightarrow$ 0.0356 (\textcolor{red!70!black}{+78.8\%}) & 0.0678 $\rightarrow$ 0.0479 (\textcolor{green!70!black}{-29.4\%}) & 0.1223 $\rightarrow$ 0.0379 (\textcolor{green!70!black}{-69.0\%}) \\
8M & 0.0149 $\rightarrow$ 0.0400 (\textcolor{red!70!black}{+167.4\%}) & 0.0432 $\rightarrow$ 0.0469 (\textcolor{red!70!black}{+8.4\%}) & 0.0959 $\rightarrow$ 0.0328 (\textcolor{green!70!black}{-65.8\%}) \\
16M & 0.0122 $\rightarrow$ 0.0314 (\textcolor{red!70!black}{+157.9\%}) & 0.0313 $\rightarrow$ 0.0385 (\textcolor{red!70!black}{+22.8\%}) & 0.0739 $\rightarrow$ 0.0298 (\textcolor{green!70!black}{-59.8\%}) \\
\bottomrule
\end{tabular}

\end{table*}

\subsection{Qualitative Results}

We present representative reconstruction results for selected test samples at the 40th and 60th percentiles of spectral error for the 16M model at encoder depth 4 in appendix ~\ref{app:samples}. Each sample is visualized as a mosaic showing ground truth, reconstructions from the model (VAE or DiffCoder), and interpolation baselines (bilinear and bicubic), corresponding error fields, bar plots for the evaluation metrics, and kinetic energy spectra.

The VAE reconstructions appear overly smooth, lacking the fine-scale structures present in the ground truth. Interpolation methods produce even blurrier results, failing to recover high-frequency content. DiffCoder reconstructions, while not matching the ground truth pointwise, exhibit realistic small-scale features that better match the visual texture of the flow.

The error fields reveal that VAE errors are spatially correlated with regions of high vorticity gradients, indicating systematic smoothing. DiffCoder errors, while sometimes larger in magnitude, are more spatially uniform, suggesting that the model generates plausible structures even where it cannot exactly reconstruct them. The energy spectra subplots quantify this: the VAE spectrum rolls off at high wavenumbers, while DiffCoder maintains spectral content closer to the ground truth across the full wavenumber range.

\section{Conclusion}
\label{sec:conclusion}

We have presented DiffCoder, a framework that couples a convolutional encoder with a conditional diffusion model as the decoder for compressing and reconstructing flow fields. By replacing the simple decoder of VAE with an iterative generative process, DiffCoder reconstructs physically plausible high-frequency content that cannot be inferred from heavily compressed representations alone.

Our experiments on two-dimensional Kolmogorov flow reveal a clear regime where diffusion-based decoding excels: under aggressive compression (encoder depth 4), DiffCoder substantially outperforms a matched VAE baseline in preserving the kinetic energy spectrum, particularly at high wavenumbers corresponding to small-scale turbulent structures. This advantage diminishes under mild compression (depth 2), where sufficient information survives the bottleneck for deterministic reconstruction.

These findings have practical implications for large-scale fluid flow simulations where storage constraints necessitate aggressive compression. While pointwise reconstruction errors may be unavoidable at extreme compression ratios, preserving statistical and spectral properties may enable a more meaningful downstream analysis of compressed data.

Several directions remain for future work. Extending DiffCoder to three-dimensional flows would test scalability to the high-dimensional settings most relevant to practical applications. The learned latent space also opens possibilities for latent-space forecasting: training temporal models (e.g., transformers or neural ODEs) to predict dynamics directly in the compressed representation, then decoding trajectories on demand. Such an approach could enable efficient long-horizon prediction while leveraging the diffusion decoder to reconstruct statistically faithful flow fields at each timestep. Additionally, incorporating physics-informed losses, such as divergence-free constraints or spectral penalties, could further improve statistical fidelity.

\section*{Data Availability Statement}

The data used for training and evaluation of the model will be publicly available upon publication.

\appendix

\section{Representative Sample Reconstructions}
\label{app:samples}

This appendix presents representative reconstruction results for selected test samples at the 40th and 60th percentiles of spectral error for the 16M model at encoder depth 4 (leading to compressed field of size $64\times64$). Each figure shows a mosaic containing ground truth, model reconstruction, bilinear and bicubic interpolation baselines, error fields, reconstruction error bar plots, and kinetic energy spectra comparison.

\begin{figure*}[!h]
\centering
\includegraphics[width=0.8\textwidth, trim={2cm 0 2cm 0}, clip]{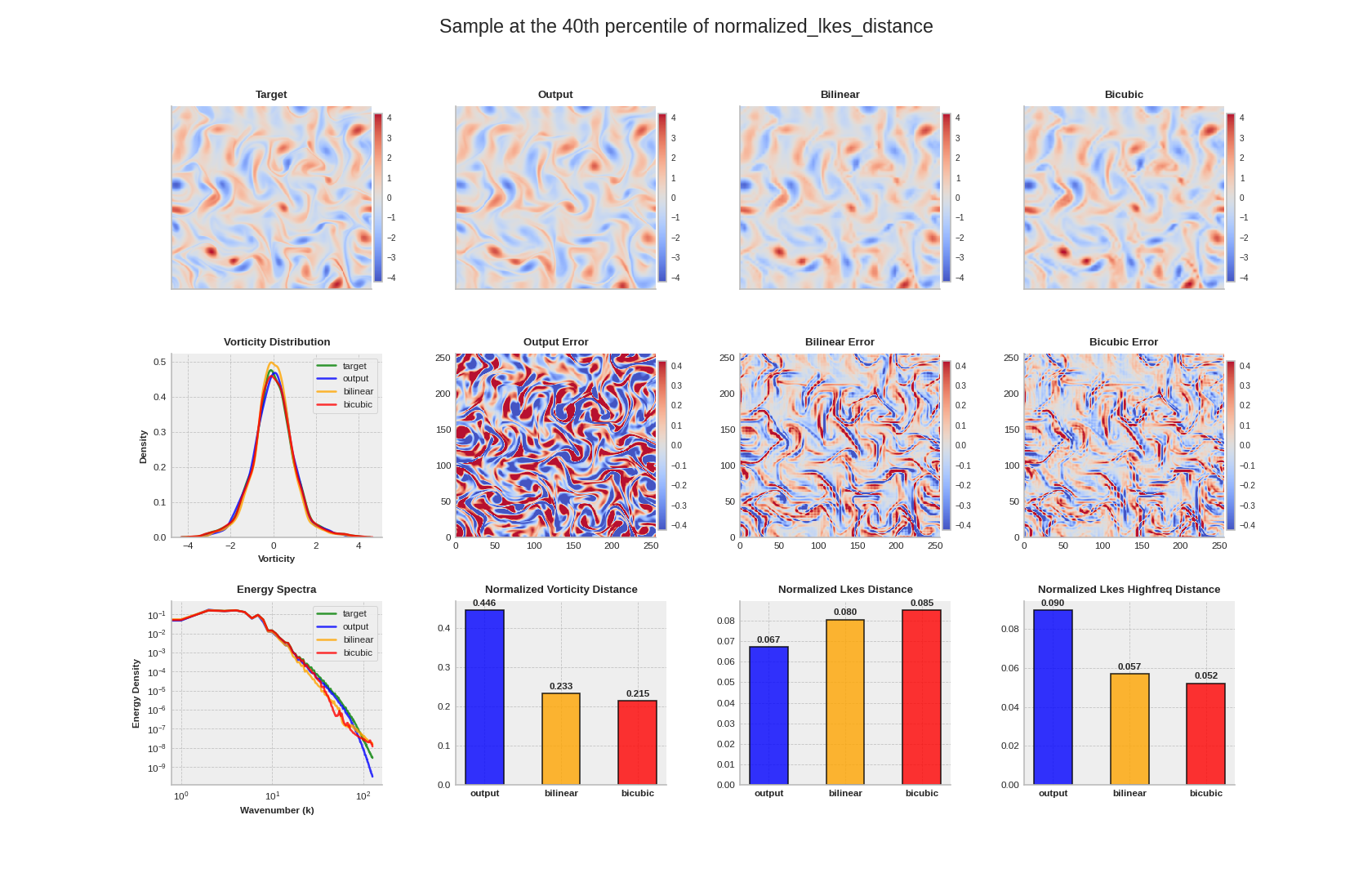}
\caption{VAE reconstruction at 40th percentile of spectral error (model size 16M, depth 4).}
\label{fig:sample_vae_p40}
\end{figure*}

\begin{figure*}[!h]
\centering
\includegraphics[width=0.8\textwidth, trim={2cm 0 2cm 0}, clip]{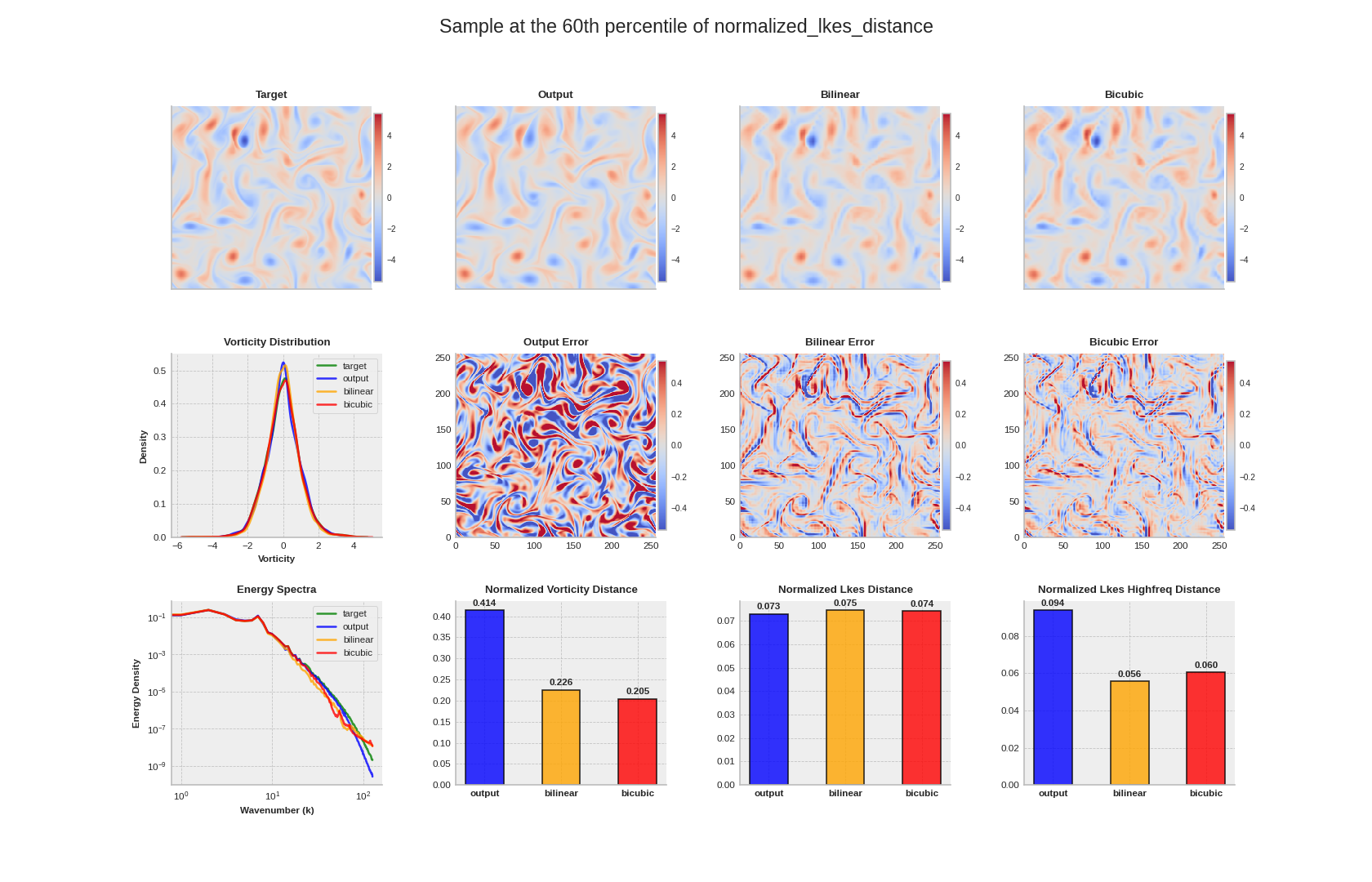}
\caption{VAE reconstruction at 60th percentile of spectral error (model size 16M, depth 4).}
\label{fig:sample_vae_p60}
\end{figure*}

\begin{figure*}[!h]
\centering
\includegraphics[width=0.8\textwidth, trim={2cm 0 2cm 0}, clip]{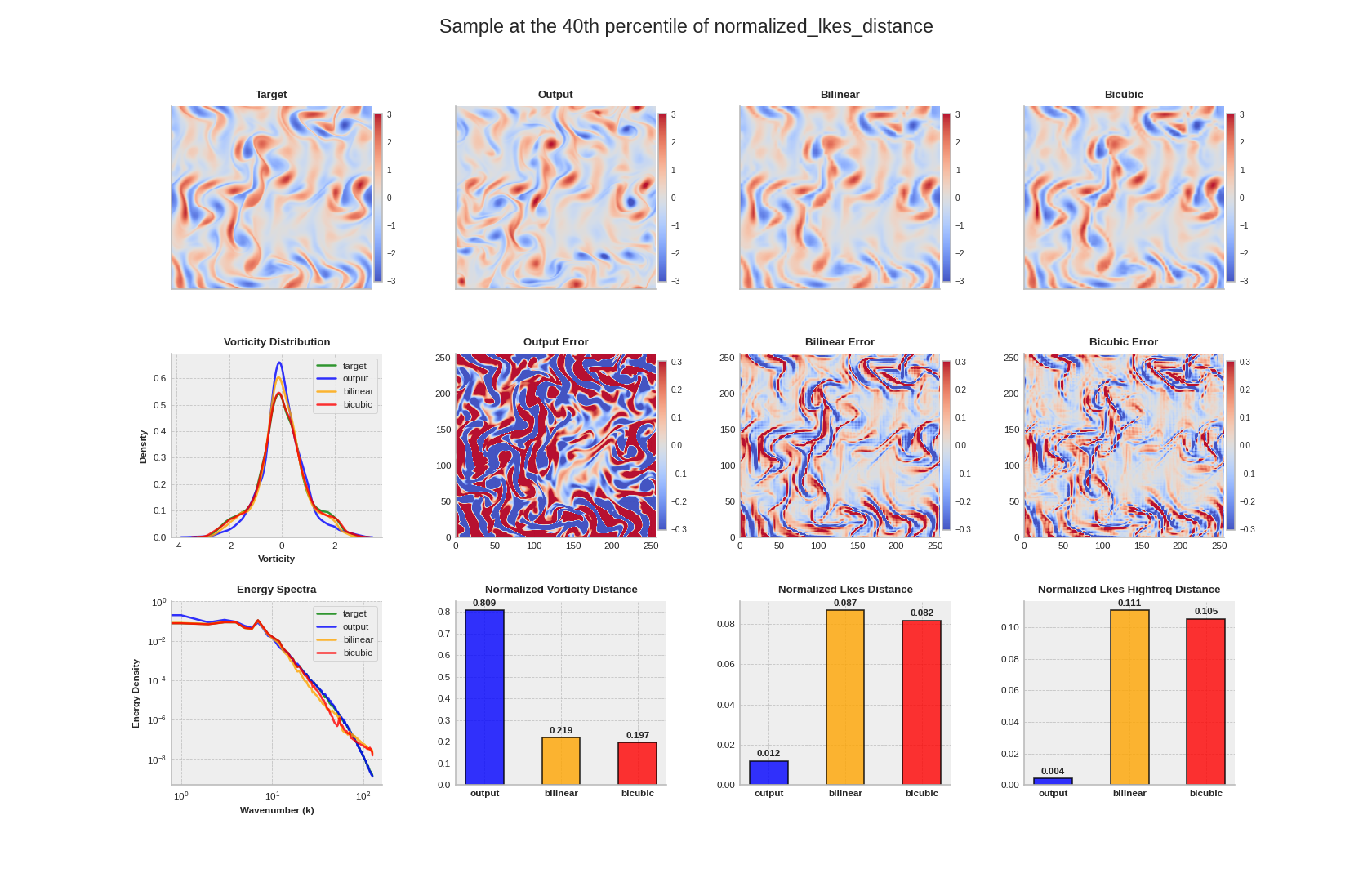}
\caption{DiffCoder reconstruction at 40th percentile of spectral error (model size 16M, depth 4).}
\label{fig:sample_diff_p40}
\end{figure*}

\begin{figure*}[!h]
\centering
\includegraphics[width=0.8\textwidth, trim={2cm 0 2cm 0}, clip]{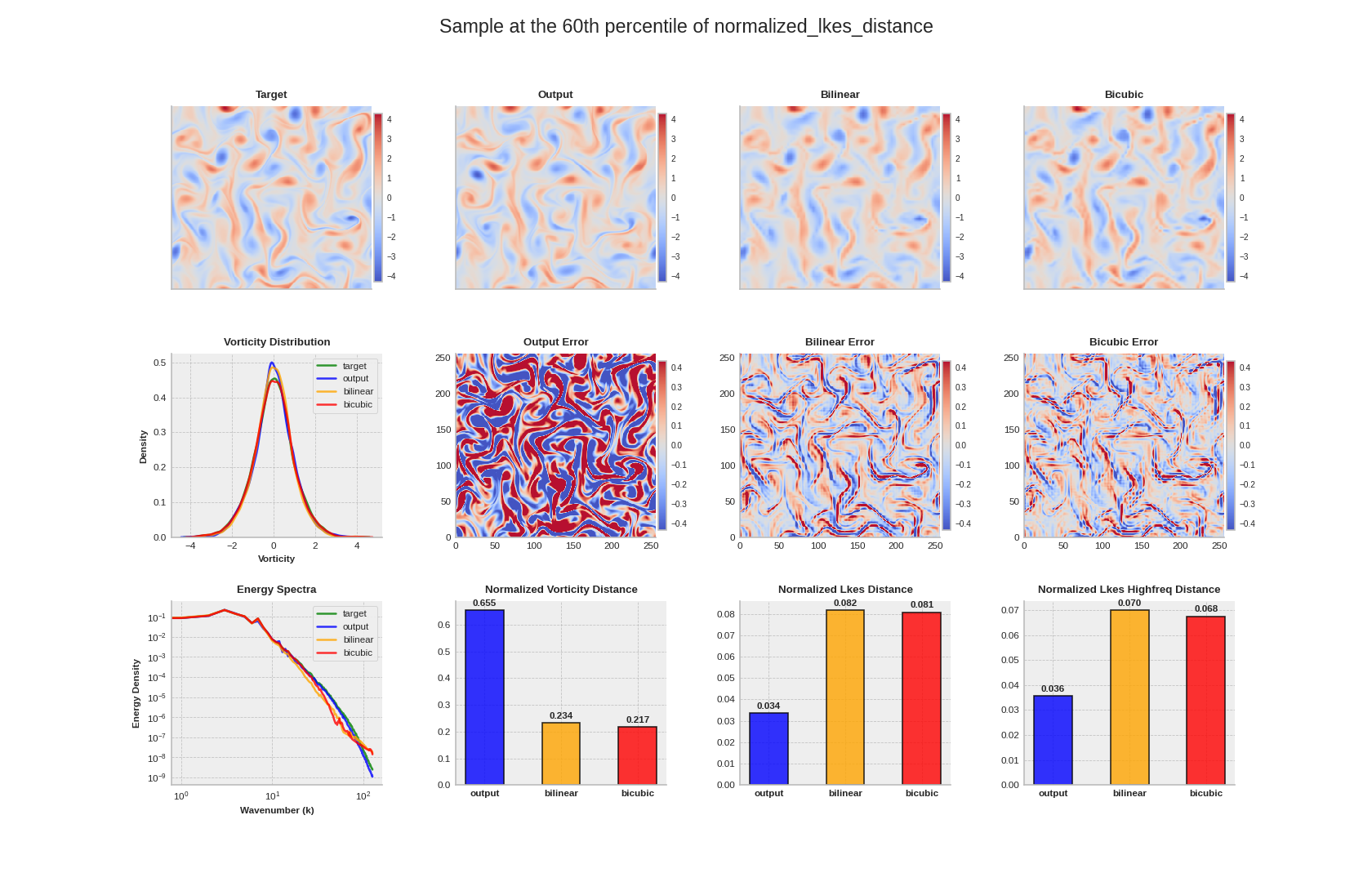}
\caption{DiffCoder reconstruction at 60th percentile of spectral error (model size 16M, depth 4).}
\label{fig:sample_diff_p60}
\end{figure*}

\clearpage
\bibliography{DiffCoder}

@article{Glaws2020DeepLearningInSitu,
  author = {Glaws, Adrian and Papas, Peter and Militzer, Matthias and Christlieb, Andrew J.},
  title = {Deep learning for in situ data compression of large turbulent flow simulations},
  journal = {Physical Review Fluids},
  year = {2020},
  volume = {5},
  number = {11},
  pages = {114602},
  doi = {10.1103/PhysRevFluids.5.114602}
}

@article{Momenifar2022AutoencoderTurbulence,
  author = {Momenifar, Mahdi and others},
  title = {A Physics-Informed Vector Quantized Autoencoder for Data Compression of Turbulent Flow},
  journal = {arXiv preprint arXiv:2201.03617},
  year = {2022},
  note = {preprint}
}

@article{Racca2023PredictingTurbulentDynamics,
  author = {Racca, Alessandro and others},
  title = {Predicting turbulent dynamics with the convolutional autoencoder echo state network},
  journal = {Journal of Fluid Mechanics},
  year = {2023},
  doi = {10.1017/jfm.2023.xxx}
}

@article{Fukami2021ModelOrderReductionNN,
  author = {Fukami, Kenta and others},
  title = {Model order reduction with neural networks: Application to laminar and turbulent flows},
  journal = {SN Applied Sciences},
  year = {2021},
  volume = {3},
  pages = {655},
  doi = {10.1007/s42979-021-00867-3}
}

@article{SoleraRico2024BetaVAETransformers,
  author = {Solera‐Rico, Alberto and others},
  title = {$\beta$-Variational autoencoders and transformers for reduced-order modelling of fluid flows},
  journal = {Nature Communications},
  year = {2024},
  volume = {15},
  pages = {45578},
  doi = {10.1038/s41467-024-45578-4}
}

@article{BayesianCondDiffusionTurbulence2023,
  author = {Anonymous, et al.},
  title = {Bayesian Conditional Diffusion Models for Versatile Spatiotemporal Turbulence Generation},
  journal = {arXiv preprint arXiv:2311.07896},
  year = {2023},
  note = {preprint}
}

@article{UnfoldingTimeGenerativeTurbulentFlows2024,
  title={Unfolding time: Generative modeling for turbulent flows in 4d},
  author={Saydemir, Abdullah and Lienen, Marten and G{\"u}nnemann, Stephan},
  journal={arXiv preprint arXiv:2406.11390},
  year={2024}
}

@article{shu2023physics,
  title={A physics-informed diffusion model for high-fidelity flow field reconstruction},
  author={Shu, Dule and Li, Zijie and Farimani, Amir Barati},
  journal={Journal of Computational Physics},
  volume={478},
  pages={111972},
  year={2023},
  publisher={Elsevier}
}

@article{jadhav2023stressd,
  title={StressD: 2D Stress estimation using denoising diffusion model},
  author={Jadhav, Yayati and Berthel, Joseph and Hu, Chunshan and Panat, Rahul and Beuth, Jack and Farimani, Amir Barati},
  journal={Computer Methods in Applied Mechanics and Engineering},
  volume={416},
  pages={116343},
  year={2023},
  publisher={Elsevier}
}

@article{shu2024zero,
  title={Zero-shot uncertainty quantification using diffusion probabilistic models},
  author={Shu, Dule and Farimani, Amir Barati},
  journal={arXiv preprint arXiv:2408.04718},
  year={2024}
}

@article{ogoke2024inexpensive,
  title={Inexpensive high fidelity melt pool models in additive manufacturing using generative deep diffusion},
  author={Ogoke, Francis and Liu, Quanliang and Ajenifujah, Olabode and Myers, Alexander and Quirarte, Guadalupe and Malen, Jonathan and Beuth, Jack and Farimani, Amir Barati},
  journal={Materials \& Design},
  volume={245},
  pages={113181},
  year={2024},
  publisher={Elsevier}
}

@article{jadhav2024generative,
  title={Generative lattice units with 3d diffusion for inverse design: Glu3d},
  author={Jadhav, Yayati and Berthel, Joeseph and Hu, Chunshan and Panat, Rahul and Beuth, Jack and Barati Farimani, Amir},
  journal={Advanced Functional Materials},
  volume={34},
  number={41},
  pages={2404165},
  year={2024},
  publisher={Wiley Online Library}
}

@article{zhou2410text2pde,
  title={Text2pde: Latent diffusion models for accessible physics simulation, 2025},
  author={Zhou, Anthony and Li, Zijie and Schneier, Michael and Buchanan Jr, John R and Farimani, Amir Barati},
  journal={URL https://arxiv. org/abs/2410.01153},
  year={2025}
}

@article{ogoke2025deep,
  title={Deep learning based optical image super-resolution via generative diffusion models for layerwise in-situ LPBF monitoring},
  author={Ogoke, Francis and Suresh, Sumesh Kalambettu and Adamczyk, Jesse and Bolintineanu, Dan and Garland, Anthony and Heiden, Michael and Farimani, Amir Barati},
  journal={Additive Manufacturing},
  pages={104790},
  year={2025},
  publisher={Elsevier}
}

@article{salimans2022progressive,
  title={Progressive distillation for fast sampling of diffusion models},
  author={Salimans, Tim and Ho, Jonathan},
  journal={arXiv preprint arXiv:2202.00512},
  year={2022}
}

@inproceedings{ronneberger2015u,
  title={U-net: Convolutional networks for biomedical image segmentation},
  author={Ronneberger, Olaf and Fischer, Philipp and Brox, Thomas},
  booktitle={International Conference on Medical image computing and computer-assisted intervention},
  pages={234--241},
  year={2015},
  organization={Springer}
}

@article{song2020denoising,
  title={Denoising diffusion implicit models},
  author={Song, Jiaming and Meng, Chenlin and Ermon, Stefano},
  journal={arXiv preprint arXiv:2010.02502},
  year={2020}
}

@article{ho2020denoising,
  title={Denoising diffusion probabilistic models},
  author={Ho, Jonathan and Jain, Ajay and Abbeel, Pieter},
  journal={Advances in neural information processing systems},
  volume={33},
  pages={6840--6851},
  year={2020}
}

@article{erichson2020shallow,
  title={Shallow neural networks for fluid flow reconstruction with limited sensors},
  author={Erichson, N Benjamin and Mathelin, Lionel and Yao, Zhewei and Brunton, Steven L and Mahoney, Michael W and Kutz, J Nathan},
  journal={Proceedings of the Royal Society A},
  volume={476},
  number={2238},
  pages={20200097},
  year={2020},
  publisher={The Royal Society Publishing}
}

@article{fukami2021global,
  title={Global field reconstruction from sparse sensors with Voronoi tessellation-assisted deep learning},
  author={Fukami, Kai and Maulik, Romit and Ramachandra, Nesar and Fukagata, Koji and Taira, Kunihiko},
  journal={Nature Machine Intelligence},
  volume={3},
  number={11},
  pages={945--951},
  year={2021},
  publisher={Nature Publishing Group UK London}
}

@article{liu2024uncertainty,
  title={Uncertainty-Aware Surrogate Models for Airfoil Flow Simulations with Denoising Diffusion Probabilistic Models},
  author={Liu, Qiang and Thuerey, Nils},
  journal={AIAA Journal},
  pages={1--22},
  year={2024},
  publisher={American Institute of Aeronautics and Astronautics}
}

@article{yang2023denoising,
  title={A denoising diffusion model for fluid field prediction},
  author={Yang, Gefan and Sommer, Stefan},
  journal={arXiv preprint arXiv:2301.11661},
  year={2023}
}

@article{li2023multi,
  title={Multi-scale reconstruction of turbulent rotating flows with generative diffusion models},
  author={Li, Tianyi and Lanotte, Alessandra S and Buzzicotti, Michele and Bonaccorso, Fabio and Biferale, Luca},
  journal={Atmosphere},
  volume={15},
  number={1},
  pages={60},
  year={2023},
  publisher={MDPI}
}

@article{bastek2024physics,
  title={Physics-Informed Diffusion Models},
  author={Bastek, Jan-Hendrik and Sun, WaiChing and Kochmann, Dennis M},
  journal={arXiv preprint arXiv:2403.14404},
  year={2024}
}

@article{shan2024pird,
  title={PiRD: Physics-informed Residual Diffusion for Flow Field Reconstruction},
  author={Shan, Siming and Wang, Pengkai and Chen, Song and Liu, Jiaxu and Xu, Chao and Cai, Shengze},
  journal={arXiv preprint arXiv:2404.08412},
  year={2024}
}

@article{sirovich1987turbulence,
  title={Turbulence and the dynamics of coherent structures. I. Coherent structures},
  author={Sirovich, Lawrence},
  journal={Quarterly of applied mathematics},
  volume={45},
  number={3},
  pages={561--571},
  year={1987}
}

@article{bui2004aerodynamic,
  title={Aerodynamic data reconstruction and inverse design using proper orthogonal decomposition},
  author={Bui-Thanh, Tan and Damodaran, Murali and Willcox, Karen},
  journal={AIAA journal},
  volume={42},
  number={8},
  pages={1505--1516},
  year={2004}
}

@article{venturi2004gappy,
  title={Gappy data and reconstruction procedures for flow past a cylinder},
  author={Venturi, Daniele and Karniadakis, George Em},
  journal={Journal of Fluid Mechanics},
  volume={519},
  pages={315--336},
  year={2004},
  publisher={Cambridge University Press}
}

@article{loiseau2018sparse,
  title={Sparse reduced-order modelling: sensor-based dynamics to full-state estimation},
  author={Loiseau, Jean-Christophe and Noack, Bernd R and Brunton, Steven L},
  journal={Journal of Fluid Mechanics},
  volume={844},
  pages={459--490},
  year={2018},
  publisher={Cambridge University Press}
}

@article{guemes2022super,
  title={Super-resolution generative adversarial networks of randomly-seeded fields},
  author={G{\"u}emes, Alejandro and Sanmiguel Vila, Carlos and Discetti, Stefano},
  journal={Nature Machine Intelligence},
  volume={4},
  number={12},
  pages={1165--1173},
  year={2022},
  publisher={Nature Publishing Group UK London}
}

@misc{super_3d_gan_1,
  title={Turbulence Enrichment using Physics-informed Generative Adversarial Networks}, 
  author={Akshay Subramaniam and Man Long Wong and Raunak D Borker and Sravya Nimmagadda and Sanjiva K Lele},
  year={2020},
  eprint={2003.01907},
  archivePrefix={arXiv},
  primaryClass={physics.comp-ph}
}

@article{super_3d_gan_2,
  title={Unsupervised deep learning for super-resolution reconstruction of turbulence},
  volume={910},
  journal={Journal of Fluid Mechanics},
  publisher={Cambridge University Press (CUP)},
  author={Kim, Hyojin and Kim, Junhyuk and Won, Sungjin and Lee, Changhoon},
  year={2021}
}

@article{dong2015image,
  title={Image super-resolution using deep convolutional networks},
  author={Dong, Chao and Loy, Chen Change and He, Kaiming and Tang, Xiaoou},
  journal={IEEE transactions on pattern analysis and machine intelligence},
  volume={38},
  number={2},
  pages={295--307},
  year={2015},
  publisher={IEEE}
}

@article{saharia2022image,
  title={Image super-resolution via iterative refinement},
  author={Saharia, Chitwan and Ho, Jonathan and Chan, William and Salimans, Tim and Fleet, David J and Norouzi, Mohammad},
  journal={IEEE transactions on pattern analysis and machine intelligence},
  volume={45},
  number={4},
  pages={4713--4726},
  year={2022},
  publisher={IEEE}
}

@inproceedings{nichol2021improved,
  title={Improved denoising diffusion probabilistic models},
  author={Nichol, Alexander Quinn and Dhariwal, Prafulla},
  booktitle={International conference on machine learning},
  pages={8162--8171},
  year={2021},
  organization={PMLR}
}

@inproceedings{hang2023efficient,
  title={Efficient diffusion training via min-snr weighting strategy},
  author={Hang, Tiankai and Gu, Shuyang and Li, Chen and Bao, Jianmin and Chen, Dong and Hu, Han and Geng, Xin and Guo, Baining},
  booktitle={Proceedings of the IEEE/CVF International Conference on Computer Vision},
  pages={7441--7451},
  year={2023}
}

@article{dhariwal2021diffusion,
  title={Diffusion models beat gans on image synthesis},
  author={Dhariwal, Prafulla and Nichol, Alexander},
  journal={Advances in neural information processing systems},
  volume={34},
  pages={8780--8794},
  year={2021}
}

@article{abaidi2025exploring,
  title={Exploring denoising diffusion models for compressible fluid field prediction},
  author={Abaidi, Rim and Adams, Nikolaus A},
  journal={Computers \& Fluids},
  pages={106665},
  year={2025},
  publisher={Elsevier}
}

@article{lu2024generative,
  title={Generative downscaling of PDE solvers with physics-guided diffusion models},
  author={Lu, Yulong and Xu, Wuzhe},
  journal={arXiv preprint arXiv:2404.05009},
  year={2024}
}

@article{chen2023importance,
  title={On the importance of noise scheduling for diffusion models},
  author={Chen, Ting},
  journal={arXiv preprint arXiv:2301.10972},
  year={2023}
}

@article{song2020score,
  title={Score-based generative modeling through stochastic differential equations},
  author={Song, Yang and Sohl-Dickstein, Jascha and Kingma, Diederik P and Kumar, Abhishek and Ermon, Stefano and Poole, Ben},
  journal={arXiv preprint arXiv:2011.13456},
  year={2020}
}

@article{li2023scalable,
  title={Scalable transformer for pde surrogate modeling},
  author={Li, Zijie and Shu, Dule and Barati Farimani, Amir},
  journal={Advances in Neural Information Processing Systems},
  volume={36},
  pages={28010--28039},
  year={2023}
}

\end{document}